# Wheeled Robots Path Planing and Tracking System Based on Monocular Visual SLAM

Ziqiang Wang, Hegen Xu, and Youwen Wan

*Abstract*—Warehouse logistics robots will work in different warehouse environments. In order to enable robots to perceive environment and plan path faster without modifying existing warehouses, we uses monocular camera to achieve an efficient robot integrated system. Mapping and path planning the two main tasks presented in this paper. The direct method visual odometry is applied to localize, and the 3D position of major obstacles in the environment is calculated. We describe the terrain with occupied grid map, the 3D points are projected onto the robot motion plane, thus accessibility of each grid is determined. Based on the terrain information, the optimized A* algorithm is used for path planning. Finally, according to localization and planning, we control the robot to track path. We also develop a path-tracking robot prototype. Simulation and experimental results verify the effectiveness and reliability of the proposed method.

## I. INTRODUCTION

In industry 4.0, the flow of goods will increase dramatically, so that each logistics warehouse must become a more intelligent and automated node in the logistics system. Using wheeled robots in warehouses can greatly improve the efficiency of warehouse goods transfer. The warehouse is a static environment, the robot should has the ability of quickly adapt to the environment, so that the robot can be rapidly deployed to different existing warehouses without transforming the warehouse.

Visual localization and mapping technology[1] was developed and used for robots during the past two decades. Camera can get more information on the environment structure than Lidar[2]. There are two types of SLAM visual odometry, which are divided into direct method and feature point method. Feature point method mainly includes PTAM[3], ORB-SLAM[19] and other algorithms.

Mur-Artal and Juan D. Tardos proposed the ORB-SLAM2, which is very easy to use. It supports monocular, binocular, and RGB-D sensors. The calculations are performed around the ORB features. ORB is not like SIFT and SURF which is time-consuming and can be calculated on the high-performance CPU in real time.

LSD-SLAM[5][8],DSO[20][21] are widely used direct methods. Jakob Engel and Daniel Cremers at the Technical University of Munich proposed a monocular SLAM algorithm based on direct method to construct a large-scale, globally consistent environment map called Large-Scale Direct (LSD) Monocular SLAM. The advantage of the feature point method is that the map drifts small and the loop-closures detection is accurate. Due to the relatively sparse feature points, it can be used for robot localization, difficult to use for robot navigation and path planning, and relatively calculate slow. Direct method can extract relatively dense environmental features for robot navigation and positioning, and can run in real time on a PC. But the map drift is large than the feature point method.

Path planning and tracking technology of robot systems are the keys to improve the efficiency. Path planning based on occupied grid map is a common method of environment for robots, such as Dijkstra[12], A*[13], D*[14] and so on. Dijkstra's main application is to find the shortest path between the start and end point of the map, but the path search is non-heuristic and slow. D* mainly solves the problem of the change of travel cost caused by the dynamic change of the environment. For a static environment such as a warehouse, the A* algorithm can find the shortest path between two points more quickly and efficiently[15][16].

The rest of the paper is organized as follows. In Section II, we present the direct visual localization and grid mapping algorithm for wheeled robot. Section III discusses path planning and robot tracking control. In Section IV, we present the simulation and experimental results. Finally, we present concluding remarks in Section V.

## II. DIRECT VISUAL LOCALIZATION AND GRIDMAPPING

### A. Localization

The direct pose tracking method[8] is based on the intensity structures of local time-varying image regions are approximately constant under motion for at least a short duration and apply in Optical Flow [4]. In current grayscale frame $I_i$, direct method select these pixels $p$ whose gradient $\partial I_i/\partial p$ is bigger to calculate. We can get sufficient obstacle texture structure.

Every time robot walks a fixed distance in route, a frame is selected as the key frame $\mathcal{K}_j$. Assume the 3D rigid body transform[6] from $\mathcal{K}_j$ to $I_i$ is $T_{ji} \in SE(3)$. $T_{ji}$ is combined by rotation matrix $T_{ji} \in SO(3)$ and translation vector $t$ from (1).

$$T_{ji} = \begin{pmatrix} R_{ji} & t_{ji} \\ 0^T & 1 \end{pmatrix}_{4\times 4} \qquad (1)$$

Corresponding Lie-algebra[6] is $\delta_{ji} \in \mathfrak{se}(3)$, $T = exp(\delta_{ji})$. Before next frame $\mathcal{K}_{j+1}$ was chosen, using Levenberg-Marquardt[7] with residual weights method to optimize $\delta_{ji}$ from (2).

Ziqiang Wang is postgraduate student with the college of Electronics and Information Engineering of Tongji Universty, Shanghai China. (e-mail: 1531651@ tongji.edu.cn).

Assume the inverse depth of $p$ is $d_p \sim N(\mu_p, \sigma_p^2)$, its initialization value is 1. $W$ is a weight that is positively related to the variance of depth, meaning that the greater the variance, the point is less reliable in optimization. $\sigma_I^2$ is the gaussian gray scale noise of the image and it is a constant value.

$$\begin{cases} \delta^* = \arg\min_{\delta}\{E_p(\delta_{ji})\}, \frac{\partial I_i}{\partial p} > T_g \\ E(\delta_{ji}) = \sum_{p \in \Omega_{D_i}} W_{r_p}(p, \delta_{ji}) r_p^2(p, \delta_{ji}) \\ r_p(p, \delta_{ji}) := I_i(p) - I_j(p, D_i(p), \delta_{ji}) \\ W_{r_p}(p, \delta_{ji}) := \left(2\sigma_I^2 + \left(\frac{\partial r_p(p, \delta_{ji})}{\partial \mu_p}\right)^2 \sigma_p^2\right)^{-1} \end{cases} \quad (2)$$

*B. Depth Estimation*

Based on optimal pose $\delta_{ji}$ of the camera, stereo matching method[5] in epipolar geometry can estimate the inverse depth mean value $\mu_p$ of the current key frame $\mathcal{K}_j$. Because camera pose estimate error, there will be geometric orientation variance $\sigma_{geo}^2$. Another is the photometric variance $\sigma_{pho}^2$ caused by the noise of the camera photosensitive. Thus $\sigma_p^2 = \sigma_{pho}^2 + \sigma_{geo}^2$.

When ordinary frame $I_i$ came, all large gradient pixel in $\mathcal{K}_j$ which also exist in $I_i$ can solve a priori normal distribution $\mathcal{N}(\mu_p, \sigma_p^2)$. Once next frame $I_{i+1}$ came, the post normal distribution $\mathcal{N}(\mu_o, \sigma_o^2)$ can be calculated. Fusion distribution will from(3).

$$\mathcal{N}\left(\frac{\sigma_p^2 d_o + \sigma_o^2 d_p}{\sigma_p^2 + \sigma_o^2}, \frac{\sigma_p^2 \sigma_o^2}{\sigma_p^2 + \sigma_o^2}\right) \quad (3)$$

When next key frame $\mathcal{K}_{j+1}$ came, optimal pose is $\delta_{jj+1}$, the inverse depth of $\mathcal{K}_j$ stop fusion and replaced by $\mathcal{K}_{j+1}$. Continuously we can get key frames set $K$ as shown in Fig.1:

$$K = \{\mathcal{K}_i | p = [u, v]^T, d_p = \mu_p, \frac{\partial \mathcal{K}_i}{\partial p} > T_g, T_{jj+1} = exp(\delta_{jj+1}), j \in \mathbb{N}\}$$

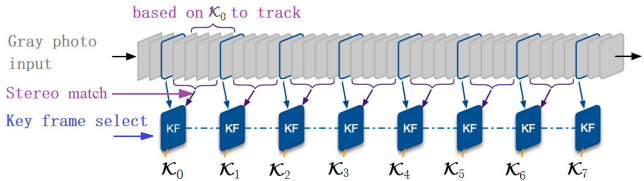

Figure 1. Continuously creat key frames.

*C. Occupied Grid Mapping*

At time $t$, according to the key frames set $K$, we can get all 3D coordinates. Considering a pixel $p = [u, v]^T$ in $\mathcal{K}_j$, its 3D coordinates $P$ can be calculated by $P = d_p^{-1} C^{-1} p$, $C$ is the camera calibration intrinsic matrix. We choose optical coordinate to first key frame $\mathcal{K}_0$ as the robot world coordinate. So each $P$ should convert to world coordinate $P_{world}$ from (4).

$$P_{world} = \left(\prod_{i=0}^{N-1} T_{j(j+1)}^{-1}\right) P$$
$$T_{j(j+1)}^{-1} = \begin{pmatrix} R_{j(j+1)}^T & -R_{j(j+1)}^T t_{j(j+1)} \\ 0^T & 1 \end{pmatrix} \quad (4)$$

Let all $P$ of all key frame convert world. We get $n$ points and form the 3D point set $A$:

$A = \{a_1 = [x_1, y_1, z_1]^T, \ldots, a_n = [x_n, y_n, z_n]^T\}$

$\forall a \in A$, let $y = 0$, projecting all points to the $OXZ$ plane. Then a new set $P$ of points is formed:

$P = \{p_1 = [x_1, 0, z_1]^T, \ldots, p_m = [x_m, 0, z_m]^T\}$

Use fixed intervals to divide the entire plane into many small grid squares with a fixed horizontal grid spacing of $H$, and vertical spacing of $V$, the grid nodes can be represented as a set $Q$ of points:

$Q = \{q = [h, v]^T | 0 \leq |h| \leq H_{sum}, 0 \leq |v| \leq V_{sum}, h, v \in \mathbb{N}\}$

There is a mapping $f: P \to Q$ between set $P$ and $Q$ from (5).

$$f := \begin{cases} X_{max} = MAX\{x_1, x_2, \ldots, x_m\} \\ X_{min} = MIN\{x_1, x_2, \ldots, x_m\} \\ H_s = \frac{X_{max} - X_{min}}{H} \\ Z_{max} = MAX\{z_1, z_2, \ldots, z_m\} \\ Z_{min} = MIN\{z_1, z_2, \ldots, z_m\} \\ V_s = \frac{Z_{max} - Z_{min}}{V} \\ p_i = [x_i, z_i]^T \in P, q_i = [h_i, v_i]^T \in Q \\ h_i = \left\lfloor \frac{x_i}{H} \right\rfloor, v_i = \left\lfloor \frac{z_i}{V} \right\rfloor \end{cases} \quad (5)$$

Finally, an Occupation Grid Map(OGM) is created. After mapping $P$ to $Q$, the interior of the grid $q_i$ will contain $q$ points. As shown in Figure 2, a threshold $T_1$ is selected. When $q > T_1$, it indicates that the grid is unreachable. Otherwise, it indicates that grid can be accessed. So that OGM maps can be made.

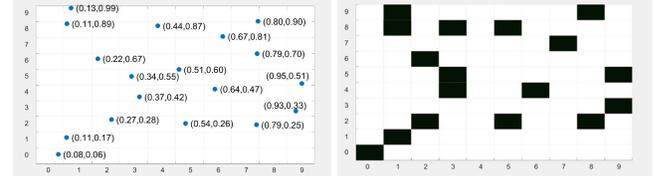

Figure 2. Creat occupation grid map.

### III. PATH PLANING AND TRACKING CONTROL

*A. Time optimized A* Algorithm*

The A* algorithm synthesizes the evaluation function of Dijkstra and Greedy algorithm. The algorithm relies on the evaluation function $f(n) = g(n) + h(n)$ to measure whether the path is optimal. The evaluation $f(n)$ is a valuation function of a grid node. $g(n)$ represents the movement cost evaluation of node n from the starting point, which is the movement distance of all the passed parent nodes by adding n to the starting point; $h(n)$ is the cost to the end point, which is a heuristic value where the Manhattan distance is used to guide the algorithm to find the end point. We implement it in Algorithm 1.

The A* algorithm's input is the starting grid node coordinates and a grid map whose size is $N = H_s \times V_s$. Traditionally only setting up two pointer linked list of grid nodes, open list and close list.

When initialize, adding the start point to the open set, find the smallest point around the start point, and add it to the closed set until finding the end point. Past practice perform three main operations on the two pointer linked lists :

- Deleting the smallest $f(n)$ node in the open set (line 7)
- Adding it to the closed set (line 8)
- Inserting the new node into the open set. (line 17)

As we all know, time complexity of inserting and deleting operation in pointer linked list is $O(N)$, sort is $O(Nlog(N))$. So total time complexity is $O(Nlog(N) + 4N))$.

We use better data structure to perform. A priority queue[9] is a data structure that allows insert, find element with time complexity $O(log(N))$, delete the minimum element with $O(1)$. Using Priority queue to store open set. A hash table[9] is a data structure that allows insert, find element with $O(1)$. Hash table store close set. Thus total time complexity is reduce to $O(3 + 2log(N))$. The final plan path is a series of grid set $G$:

$$G = \{g = [x,z]^T | 0 \leq |x| \leq H_s, 0 \leq |z| \leq V_s, x,z \in \mathbb{N}, g \in Q\}$$

**Algorithm 1** : Time optimized A* Algorithm
**Require:** $Start, End, Q$
**Ensure:** Father node of $End$
1: function ASTARSEARCH($Start, End, Q$)
2:    $open$ = binary heap containing $Start$ node
3:    $closed$ = empty set
4:    $movecost(x,y)$ = distance from node $x$ to node $y$
5:    while $End$ node not in $open$ do
6:      $i$ = node with min $f(i)$ in $open$
7:      remove $i$ from $open$
8:      add $i$ to $closed$
9:      $count = 0$
10:      for $j$ = neighbor node of $i$ and not in $closed$ and reachable in $Q$ do
11:        $count++$
12:        $cost = g(i) + movecost(i,j)$
13:        if $j$ in $open$ and $cost < g(j)$ then
14:          remove $j$ from $open$
15:        end if
16:        if $j$ not in $open$ and not in $closed$ then
17:          add $j$ into $open$
18:          $f(j) = g(j) + h(j)$
19:          set father node of $j$ is $i$
20:        end if
21:      end for
22:      if $count == 0$ then
23:        can't find path
24:        break out
25:      end if
26:    end while
27: end function

### B. Robot Orientation and Velocity Control

We use Three-Wheeled Omni-Directional Mobile Robots (TOMRs) mechanical platforms. Fig. 3 left side shows the bottom view of the mobile robot, a metal disc evenly distributes three motors. Fig. 3 right side shows top view of the schematic of the robot kinematics. We will put the camera on the center of the metal disc.

We mount the camera on the center of the metal disc. The camera's optical axis coincides with the axis of motor 3 and $OXZ$ plane parallel to the ground plane, camera optical center is origin.

Robot global pose is $P_r = [x_r, z_r, \theta]^T$. Robot translation speed on the ground is $V_r$, counterclockwise rotation speed $\omega_r > 0$, The velocity vector of three wheels is $V_1, V_2, V_3$, the robotic kinematic model is obtained as (6)

$$\begin{cases} V_x = \dot{X}_r, V_z = \dot{Z}_r \\ V_r = V_x + V_z, |V_r|\cos\theta = |V_x|, |V_r|\sin\theta = |V_z| \\ \omega_r = \dot{\theta} \end{cases}$$

$$\begin{pmatrix} |V_1| \\ |V_2| \\ |V_3| \end{pmatrix} = \begin{pmatrix} -\frac{1}{2} & \frac{\sqrt{3}}{2} & L \\ -\frac{1}{2} & -\frac{\sqrt{3}}{2} & L \\ 1 & 0 & L \end{pmatrix} \begin{pmatrix} V_x \\ V_z \\ \omega_r \end{pmatrix}$$

(6)

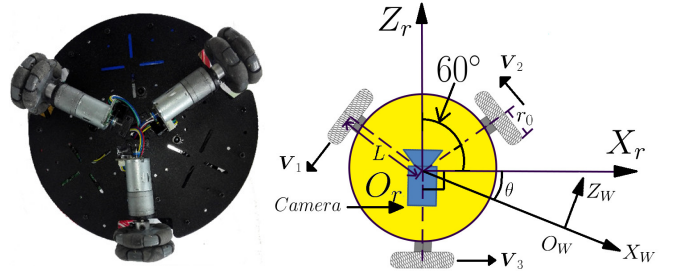

Figure 3.  Bottom view and kinematics of the robot.

### C. Track Strategies

At time $t$, we have set $K$ which contain $N$ neighboring key frames, pose between $\mathcal{K}_j$ and $\mathcal{K}_{j+1}$ is $T_{jj+1}$. 3D pose between current frame $\mathcal{I}_n$ and current key frame $\mathcal{K}_m$ is $T_{mn}$. Firstly, we can get robot pose from (7).

$$O_r = \begin{pmatrix} x_r \\ y_r \\ z_r \\ 1 \end{pmatrix} = AT0^T = \left(\prod_{i=0}^{N-1} T_{j(j+1)}^{-1}\right) T_{mn}^{-1} \begin{pmatrix} 0 \\ 0 \\ 0 \\ 1 \end{pmatrix}$$

(7)

$$n = [0,0,1,0]^T, N_r = ATn^T, \theta = \arcsin\frac{z_{N_r}}{|N_r|} \in (-\pi, \pi]$$

Three-wheel speed feedback can also be measured in real-time via encoders on wheels. Inverse operation on formula (6), we can get robot states feedback $\dot{P}_r = [\dot{x}_r, \dot{z}_r, \dot{\theta}]^T$, we consider that robot observed state in time satisfies the Markov assumption. The robot system belongs to a linear Gaussian system, so that the linear kalman filter[10] is used to estimate the robot's motion state.

We take $x_t = [x_r, \dot{x}_r, z_r, \dot{z}_r, \theta, \dot{\theta}]^T$ as the state vector of the robot system, and then measured the state of the system. State transfer equation is (8):

$$B = \begin{pmatrix} 1 & T \\ 0 & 1 \end{pmatrix}, C = \begin{pmatrix} \frac{1}{2}T^2 \\ T \end{pmatrix}$$

$$x_t = Ax_{t-1} + \varepsilon_t$$

$$= \begin{pmatrix} B & 0 & 0 \\ 0 & B & 0 \\ 0 & 0 & B \end{pmatrix} x_{t-1} + \begin{pmatrix} C & 0 & 0 \\ 0 & C & 0 \\ 0 & 0 & C \end{pmatrix} \begin{pmatrix} \ddot{x}_{t-1} \\ \ddot{z}_{t-1} \\ \ddot{\theta}_{t-1} \end{pmatrix}$$

(8)

The sampling interval is $T$ which is very short. It is the reciprocal of the frame rate of the camera. State noise vector is $\varepsilon_t$, which is six dimensions Gaussian noise vector. Acceleration is a state order higher than position and speed. We consider it is Gaussian distribution[11].

The $z_t = [x_r, \dot{x}_r, z_r, \dot{z}_r, \theta, \dot{\theta}]^T$ is the state observation vector of the system, state observation equation is :

$$z_t = Ex_{t-1} + \delta_t,$$

$E$ is a unit matrix. $\delta_t$ is state observation noise vector. Before using linear kalman filter, we determine covariance matrix $R_t$ of $\varepsilon_t$ from (9). $k_1^2, k_2^2, k_3^2$ is experience value to be present.

The covariance matrix $\boldsymbol{R}_t$ of $\boldsymbol{\delta}_t$ can be calculated with real sampling data. We get estimation sate $\boldsymbol{\mu}_t$ of $\boldsymbol{x}_t$ after update step of filter from Algorithm 2.

We get filter current time robot pose $\boldsymbol{P}_t = [x_r, z_r, \theta]^T$ and current node robot in. We set current node as start node to plan path which is a series of grid set $G$. We take two step to control according distance from current location $[x_r, z_r]^T$ to next grid center $O_i$.

$$\ddot{x}_r \sim \mathcal{N}(0, k_1^2), \ddot{z}_r \sim \mathcal{N}(0, k_2^2), \ddot{\theta}_r \sim \mathcal{N}(0, k_3^2),$$

$$\boldsymbol{D} = \begin{pmatrix} \frac{1}{4}T^4 k_1^2 & \frac{1}{2}T^3 k_1^2 \\ \frac{1}{2}T^3 k_1^2 & T^2 k_1^2 \end{pmatrix}$$

$$\boldsymbol{R}_t = \begin{pmatrix} cov(\varepsilon_1, \varepsilon_1) & \cdots & cov(\varepsilon_1, \varepsilon_6) \\ \vdots & \ddots & \vdots \\ cov(\varepsilon_6, \varepsilon_1) & \cdots & cov(\varepsilon_6, \varepsilon_6) \end{pmatrix} \quad (9)$$

$$= \begin{pmatrix} \boldsymbol{D} & 0 & 0 \\ 0 & \boldsymbol{D} & 0 \\ 0 & 0 & \boldsymbol{D} \end{pmatrix}$$

---
**Algorithm 2** : Linear Kalman Filter
**Require:** $\boldsymbol{\mu}_{t-1}, \boldsymbol{\Sigma}_{t-1}, \boldsymbol{z}_t$
**Ensure:** $\boldsymbol{\mu}_t, \boldsymbol{\Sigma}_t$
1: **function** FILTER($\boldsymbol{\mu}_{t-1}, \boldsymbol{\Sigma}_{t-1}, \boldsymbol{z}_t$)
2:    predict    $\overline{\boldsymbol{\mu}}_t = \boldsymbol{A}\boldsymbol{\mu}_{t-1}$
3:                  $\overline{\boldsymbol{\Sigma}}_t = \boldsymbol{A}\boldsymbol{\Sigma}_{t-1}\boldsymbol{A}^T + \boldsymbol{R}_t$
4:    update    $\boldsymbol{K}_t = \overline{\boldsymbol{\Sigma}}_t(\overline{\boldsymbol{\Sigma}}_t + \boldsymbol{Q}_t)^{-1}$
5:                  $\boldsymbol{\mu}_t = \overline{\boldsymbol{\mu}}_t + \boldsymbol{K}_t(\boldsymbol{z}_t - \overline{\boldsymbol{\mu}}_t)$
6:                  $\boldsymbol{\Sigma}_t = (\boldsymbol{E} - \boldsymbol{K}_t)\overline{\boldsymbol{\Sigma}}_t$
7: **end function**

---

When robot are far away from the next grid. As show in Fig. 4 left side, red point $O_r$ is location of robot, there will be an error vector $[e_x, e_y, e_\theta]^T$ when robot run. We set $V_r$ towards $O_i$ which is $\overrightarrow{O_r O_i}$, then decomposing $k_{p1}(\overrightarrow{O_r O_i}/|\overrightarrow{O_r O_i}|)$ to robot coordinate makes the robot run at a constant speed show as middle side in Fig. 4. Robot rotation is $\omega_r = k_{p2} e_\theta$.

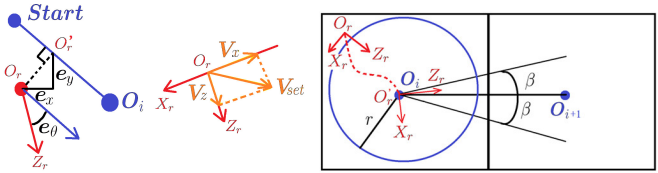

Figure 4. Control strategy of robot.

As Fig. 4 right show, red point When robot enters circle which center grid $O_i$, radius $r$. We think the robot has reached, control robot rotate towards next grid $O_{i+1}$ with $\omega_r$ and allowable error is in the range $[-\beta, \beta]$. Then robot run towards next grid $O_{i+1}$. In this way, the robot can reach its destination.

## IV. SIMULATION AND EXPERIMENT RESULTS

Considering the practical implementation, we develop an integrated robot system on physical prototype(right) and overview electrical connection diagram(left) as shown in Fig. 5.

Each 60mm omni-directional wheel is driven by a 24V DC motor. We use STM32F103 MCU developing motor driver and open UART and CAN bus interface to main computer. Industrial computer(Celeron@2.0GHz) for image progressing(global shutter, 140°, 60fps) and obstacle avoidance calculation. Lidar(6m radius, 5.5Hz data frequency)is also installed.

Robot system has four thread, which is localizing, grid mapping, controlling and user interface thread as shown in Fig. 6. The user interface (UI) is developed on OpenGL.

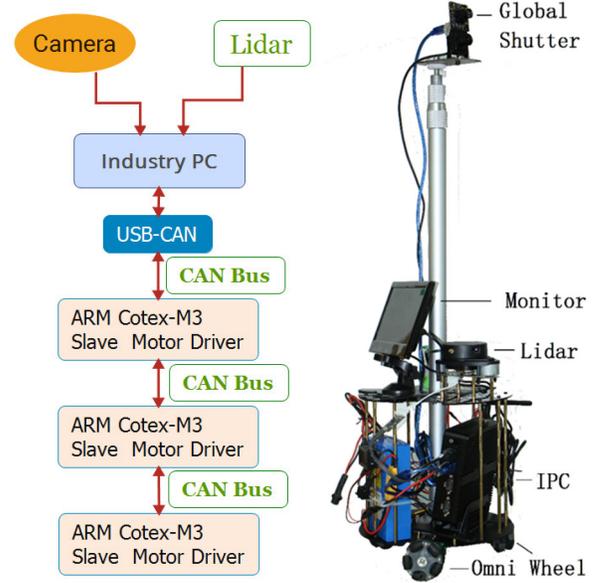

Figure 5. Overview electrical structure and physical prototype of robot.

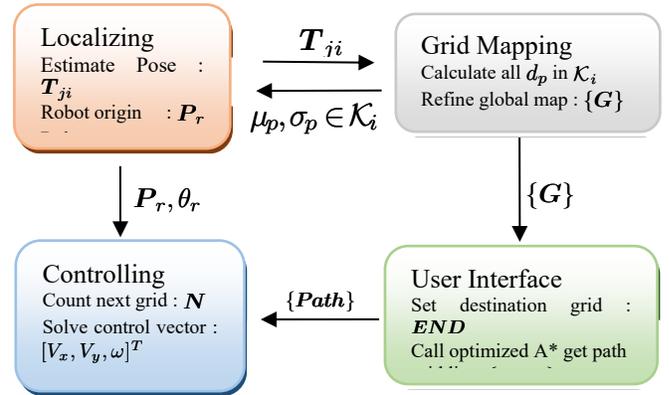

Figure 6. Overview of robot thread system.

We present four experiments to analysis the robot performance.

### A. Scale Drift Accuracy

Considering the scale of monocular vision is uncertain, there is a proportional coefficient $k_s$ compared with the scale of the real world. Each time restart the camera to build a map, we will get a different proportional coefficient $k_s$, subjecting to Gaussian distribution. Estimating $k_s$ can determine the accuracy of the map comparing with the actual geometry map.

We observe $k_s$ accurately from two directions of motion alone, let robot track the axis $O_W X_W$ and $O_W Z_W$ with fixed distance(0.2m, 0.4m, 0.6m, 0.8m, 1.0m, 1.2m, 1.4m) and record corresponding value calculated by localization$O_W Z_W$ (Vertical direction) $k_{sv} = 0.2628$, that means 0.2628 in camera is 1 meter in real world, and in $O_W X_W$ ( horizontal direction) $k_{sh} = 0.2921$ as shown in Fig. 7.

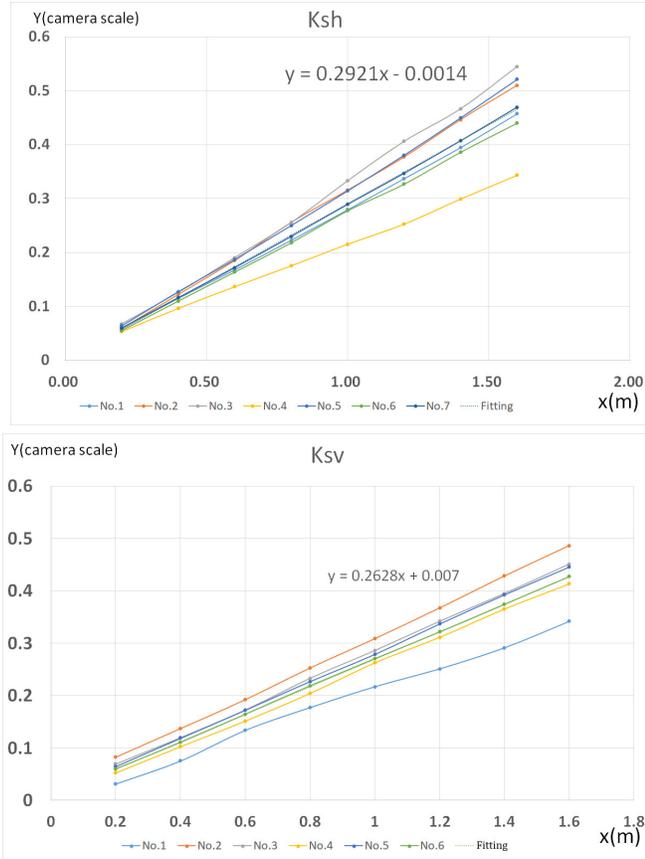

Figure 7. Scale calibration both in horizontal and Vertical.

### B. Occupied Grid Mapping

According to the formula(5), setting the fixed horizontal grid spacing and vertical spacing $H = V = 0.02$, according to $k_{sh} = 0.2921$ and $k_{sv} = 0.2628$, a grid in map is $0.076 \times 0.0685 m^2$ in real world. Grid map build threshold is $T_1 = 200$.

As shown in Fig. 8, we build occupancy grid maps in two scenes in Tonji University, upper side is $8.365 \times 7.076 m^2$ room1, down side is $6.398 \times 10.991 m^2$ room2, left side is real photos compare to right side occupancy grid maps displayed by user interface.

Headings, or heads, are organizational devices that guide the reader through your paper. There are two types: component heads and text heads.

In Tables Ⅰ, we give some attributes of two scenes, 3D points quantities, maximum and minimum grids which are integer from formula(5), both in horizontal and vertical direction of two maps. Also, we can get the error between map and real world. Because of scale uncertainty of monocular vision, the error is unsatisfactory. But error ration of length/width is very small. That means map built can achieve true proportion of the world.

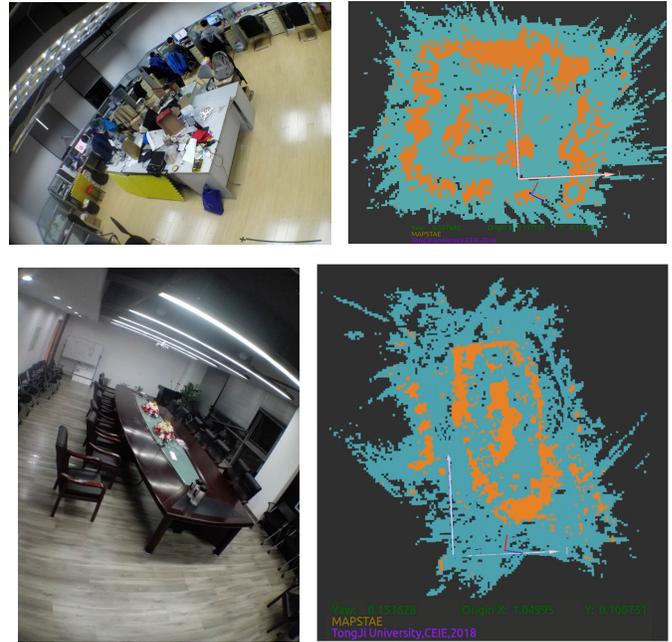

Figure 8. Map construction renderings, Above is Room1, below is Room2. Left is real world, right is map built displayed by UI, yellow is grid unreachable, blue is grid reachable. We also can see the some measure state of robot in UI. Robot origin is the red and blue axes.

### C. A* Algorithm Simulation

To evaluate the performance and performance of the A* algorithm, the algorithm was tested in the simulation software(UI). We first tests the performance of two data structure A* algorithms at different start node and end node on different size maps of 168*120, 336*240, 672*480, as shown Fig.9.

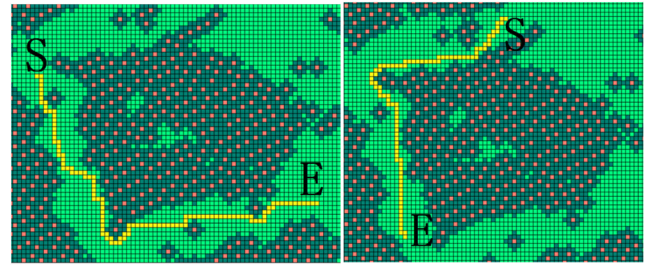

Figure 9. A* Algorithm Simulation. The yellow grid indicates the path to the plan. The letter S indicates the start node and the letter E indicates the end node.

As shown in Fig.10, the comparison of the three methods at different fifty pairs of start node and end node. The vertical axis shows the average time cost of fifty pairs. Unit is second. We use binary heap to complete priority queue. It can be seen that the time cost of A* binary heap is about 1/5 of the A* list, while the Dijkstra algorithm has the longest time, ten times as much as the A* list.

### D. Linear Kalman Filter Simulation

We implement Kalman Filter simulation based on the part of real robot pose data. We sample wheel speed at 1000Hz, then convert to $\dot{\theta}_r$ and fusion with $\theta_r$ (yaw) got by localization. According to experience, set $\ddot{\theta}_r \sim \mathcal{N}(0, 0.1^2)$, Calculating $\boldsymbol{Q}_t$ based on statistical data sampled. We sample

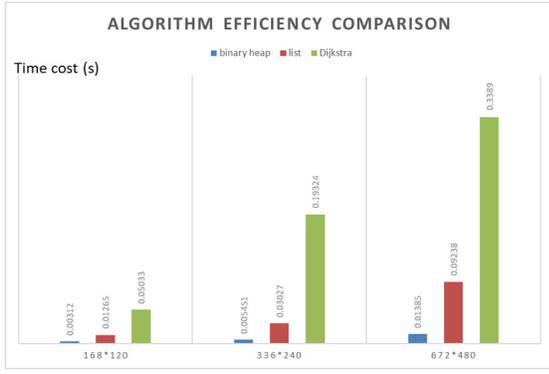

Figure 10. Vertical axis is average of fifty different pairs of start node and end node summary time cost. Horizontal axis is three different maps. Orange is time cost of A* using binary heap data structure, yellow is time cost of A* using list data structure, green is time cost of Dijkstra algorithm.

about 3500 sets of data and determine $Q_t$ from(10) and filter data in Fig.11.

$$Q_t = \begin{pmatrix} cov[\theta, \theta] & cov[\theta, \dot{\theta}] \\ cov[\dot{\theta}, \theta] & cov[\dot{\theta}, \dot{\theta}] \end{pmatrix} = \begin{pmatrix} 1023.684 & 0.221 \\ 0.221 & 25.228 \end{pmatrix} \quad (10)$$

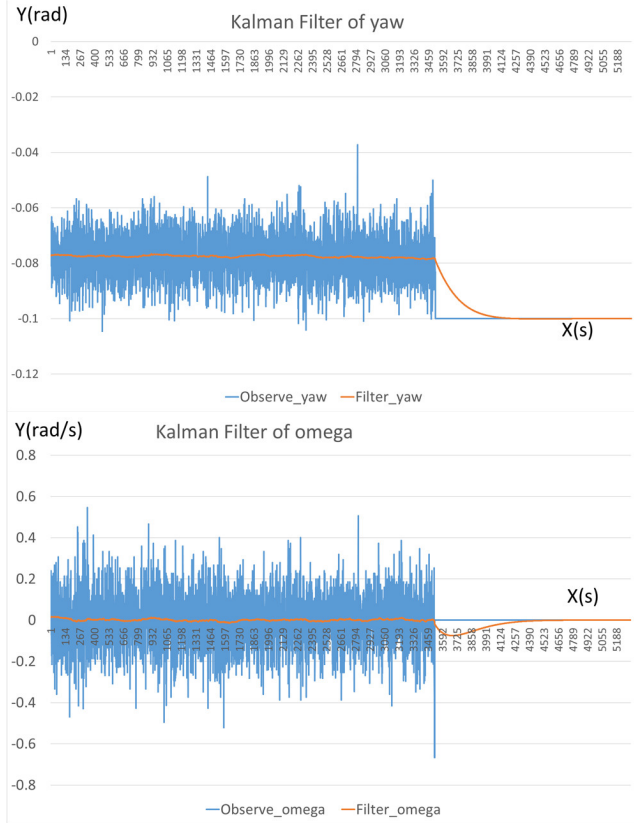

Figure 11. Yellow line is filter value, bule is measured value. It can be seen that the measured value jump is very large, and the intermediate filter value obtained is stable $\theta_r$ and $\dot{\theta}_r$ value. When only static, $\dot{\theta}_r$ value obtained is near 0, then we add simulation steering value, Filters quickly correct and converge to -0.1 rad at about 0.3s.

### D. Path Tracking

Manually set the end points of the two scenes in the UI, and then call the A* algorithm to plan path. At the same time record the coordinates of the robot's desired position $\boldsymbol{P}_e = [x_e, z_e]^T$ and current position $\boldsymbol{P}_r = [x_r, z_r]^T$.

The sampling frequency is the frame rate of the camera $F = 60Hz$, as shown in Fig.11, the small window is path planned displayed in UI.

The root mean square errors of horizontal, vertical and trajectory tracking can be calculated from (11). The results are transformed to real world with $k_{sv}$ and $k_{sh}$ shown in Table Ⅱ.

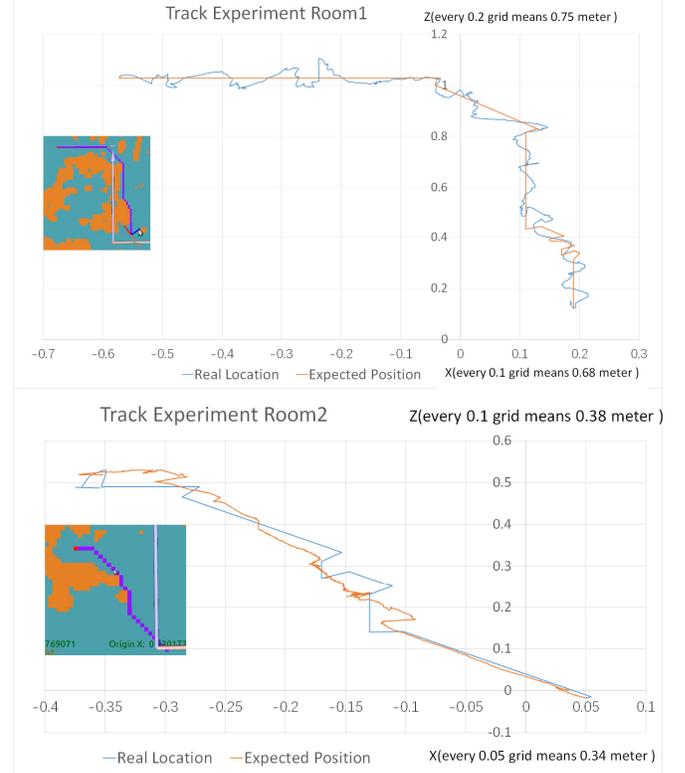

Figure 12. Yellow line is expected position, bule is real location. The coordinate is camera scale.

$$\begin{cases} RMSE(x) := \sqrt{\frac{1}{n}\sum_{i=1}^{N}(x_{r_i} - x_{e_i})^2}, RMSE(z) := \sqrt{\frac{1}{n}\sum_{i=1}^{N}(z_{r_i} - z_{e_i})^2} \\ RMSE_{track} := \sqrt{\frac{1}{n}\sum_{i=1}^{N}|\boldsymbol{P}_{r_i} - \boldsymbol{P}_{e_i}|^2} \end{cases} \quad (11)$$

TABLE I. MAP BUILT ATTRIBUTES

| Scenes | Attributes | | | |
|---|---|---|---|---|
| | *Xmax,Xmin* | *Zmax,Zmin* | *Build Map: length\*width (m\*m),* | *Build Map: ration of length/width Subhead* |
| Room1 | (70,-59) | (91,-31) | 9.804*8.35 | 1.174 |
| Room2 | (65,-44) | (142,-37) | 8.284* 13.604 | 0.609 |
| | Attributes | | | |
| | *Real: length\*width (m\*m)* | *Real: ration of length/width* | *horizon error* | *vertical error* / *Ration error* |
| Room1 | 8.365*7.076 | 1.1821 | 17.202% | 18.103% / 0.685% |
| Room2 | 6.398* 10.911 | 0.5863 | 29.477% | 24.681% / 3.8% |

TABLE II. RMSE OF TRACK

| Scenes | RMSE(x) m | RMSE(z) m | RMSE(track) m |
|---|---|---|---|
| Room1 | 0.0327245 | 0.04414 | 0.0417 |
| Room2 | 0.05754 | 0.06531 | 0.083265 |

## V. CONCLUSION

In this paper, a grid map preparation method and path planning strategy based on monocular vision simultaneous positioning and map construction are studied. Then we develop trajectory tracking strategy.

The environment is reconstructed using the direct method model, which enables the environment to be accurately modeled in various scenarios of robots. Based on the environment three-dimensional data, occupancy grid map (OGM) was prepared for the description of the terrain. According to the terrain information, A * algorithm was used to plan path, and the performance of the algorithm was improved. At last we control robot track along the path planned.

From experiments we can see the map is not accurate enough because of scale uncertainty of monocular vision. Next research will focus on using stereo vision to get accurate reconstruction. Lidar is also accurate enough, direct method combined with Lidar data fusion can be researched too.

## APPENDIX

All of our software code , hardware schematic are opened on GitHub(https://github.com/ArmstrongWall), and our demo video can be found on :

https://www.youtube.com/playlist?list=PLnJ8pi4MhtBDZLO171ndBxtIFhVsDpBkA